\documentclass[10pt,twocolumn,letterpaper]{article}

\usepackage{cvpr}              
\definecolor{cvprblue}{rgb}{0.21,0.49,0.74}

\usepackage[pagebackref,breaklinks,colorlinks,allcolors=cvprblue]{hyperref}
\usepackage{array}
\usepackage{multirow}
\usepackage[table]{xcolor}

\usepackage{array}

\newcolumntype{L}[1]{>{\raggedright\arraybackslash}p{#1}}
\newcolumntype{C}[1]{>{\centering\arraybackslash}p{#1}}

\definecolor{baselinecolor}{gray}{.9}

\newlength\savewidth

\newcolumntype{x}[1]{>{\centering\arraybackslash}p{#1pt}}
\newcolumntype{y}[1]{>{\raggedright\arraybackslash}p{#1pt}}
\newcolumntype{z}[1]{>{\raggedleft\arraybackslash}p{#1pt}}

\title{EDAR: Learning Environment-Dependent Action Representations for Robotic Manipulation}

\author{
Yuecheng Xu$^1$\textsuperscript{\dag},
Tong Yang$^{3,2}$\textsuperscript{\dag},
Jingkai Jia$^1$,
Chi Zhang$^3$, \\
Xuelong Li$^3$\textsuperscript{\ddag},
Wenqiang Zhang$^{1,2}$\textsuperscript{\ddag}\\
$^1$College of Intelligent Robotics and Advanced Manufacturing, Fudan University\\
$^2$Shanghai Key Lab of Intelligent Information Processing,\\
College of Computer Science and Artificial Intelligence, Fudan University\\
$^3$TeleAI, China Telecom\\
{\tt\footnotesize \{ycxu25, tongyang23\}@m.fudan.edu.cn, wqzhang@fudan.edu.cn}\\[-0.2em]
}

\begin{document}
\maketitle
\begingroup
\renewcommand{\thefootnote}{}
\footnotetext{\textsuperscript{\dag}Equal contribution. \quad
\textsuperscript{\ddag}Corresponding authors.}
\endgroup

\begin{abstract}
Learning effective action representations is critical for robotic manipulation, where raw control trajectories are often noisy, redundant, and difficult to model directly. Existing methods mainly encode the structure of the action stream itself, treating the role of actions in the environment as implicit. Yet manipulation is about changing the world: the same action segment can induce different outcomes under different scene contexts, making action semantics inherently environment-dependent. We propose EDAR, an Environment-Dependent Action Representation that grounds action tokens in both executable control structure and expected visual consequences. By coupling motor commands with their environment-conditioned effects, EDAR encourages the learned action space to capture interaction semantics rather than merely command-level patterns. Experiments on simulated and real-robot manipulation benchmarks demonstrate that EDAR improves downstream policy learning, especially in long-horizon manipulation. These results highlight the importance of grounding action representations in executable control structure and environment-conditioned visual change.

\end{abstract}    
\section{Introduction}
\label{sec:intro}
Raw action trajectories, especially those collected from demonstrations or teleoperation, are often noisy, redundant, and uneven. They contain high-frequency jitter, task-irrelevant micro-corrections, and embodiment-specific control patterns, making them difficult to model directly as a reliable interface for policy learning. This issue becomes more pronounced in manipulation, where policies must generate temporally coherent and physically feasible behaviors over long horizons. As a result, learning effective action representations has become an important problem in robotic manipulation. A good action representation should provide a compact and learnable interface between perception and control, while preserving the information necessary for accurate and robust task execution. The central question is therefore: what kind of action representation is appropriate for robotic manipulation?

\begin{figure*}[t]
    \centering
    \includegraphics[width=0.95\linewidth]{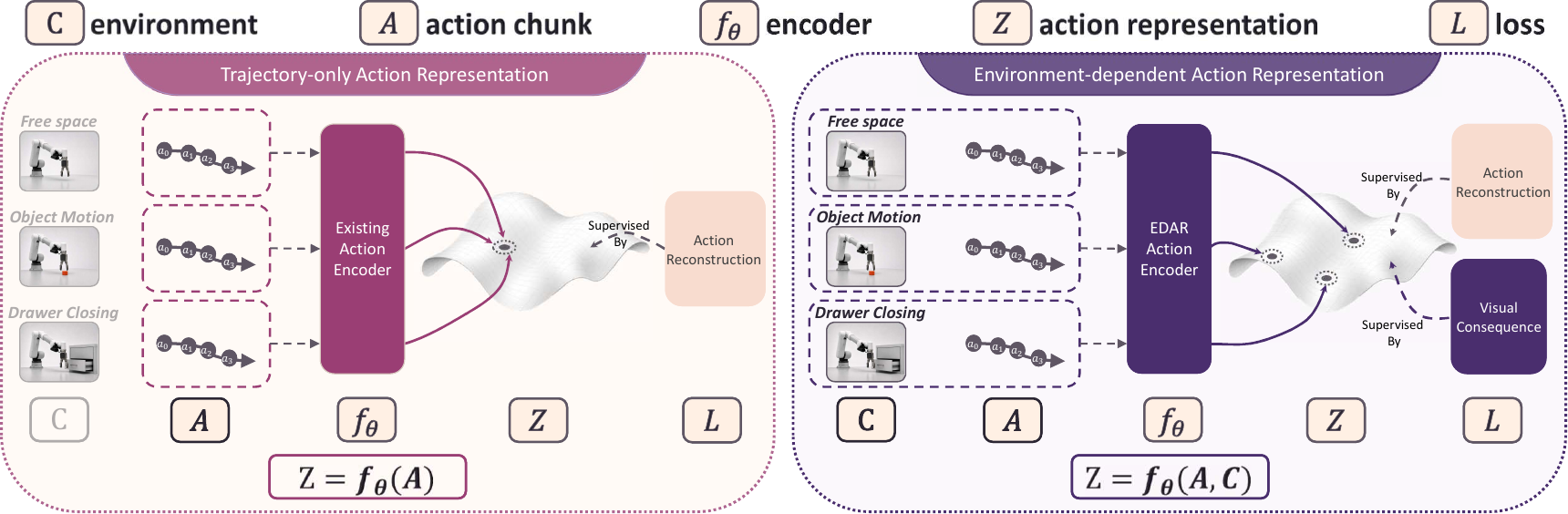}
    \vspace{-0.6em}
    \caption{\textbf{Conceptual comparison.}
    The same action segment can induce different effects under different scene contexts, and should therefore not be represented as a context-invariant representation.
    \textbf{Left:} trajectory-only action representation collapses these environment-dependent effects into the same latent action.
    \textbf{Right:} EDAR conditions the action representation on the current scene and grounds it with visual consequences, separating actions according to their context-dependent effects.}
    \label{fig:motivation}
    \vspace{-0.7em}
\end{figure*}

Most existing action representations~\cite{pertsch2025fast,wang2026latentvla} are learned primarily from the action stream itself. They may differ in form, ranging from compact trajectory codes to action tokens or frequency-based representations, but they typically share a trajectory-structure-based view: the goal is to preserve or regularize the organization of the control trajectory, such as its geometry, temporal variation, smoothness, or predictability. This view is useful for reducing redundancy and making actions easier to model, but it remains largely action-only. It describes how a command sequence is structured, while leaving its role in the environment implicit. This is limiting for manipulation, where an action is not merely a motor signal but a means of changing the world. The same gripper-closing and rightward-motion segment may be inconsequential in free space, grasp an object under contact, or close a drawer when aligned with a handle, as shown in Fig.~\ref{fig:motivation}. Such cases suggest that action semantics are not determined by the command alone, but by the environment-conditioned transition it induces. Therefore, an appropriate action representation for manipulation should not only encode the execution pattern of a command, but also be grounded in its expected effect on the scene.

In this work, we argue that learning environment-dependent action representations requires more than just feeding visual context into an action encoder. Under an action-reconstruction objective alone, the model has a shortcut: it can encode sufficient trajectory information without relying on the visual condition. As a result, the latent space remains organized primarily by command-level similarity: two action segments with similar motor patterns may be mapped nearby even when they play different roles in different scenes. To prevent this collapse, the representation must be constrained by what the action does to the environment.
We therefore introduce EDAR, an Environment-Dependent Action Representation that learns latent action tokens from both execution and effect. EDAR preserves the original control trajectory through action reconstruction, while simultaneously requiring the same tokens to model the visual consequence of executing the action under the current scene context. By letting action tokens interact with image tokens in a shared token space, EDAR allows scene context and predicted environmental change to shape the action embedding itself, rather than serving as external conditioning signals. This yields an action representation that is not merely compact, but grounded in interaction semantics.

We conduct extensive experiments in both simulated and real-world manipulation settings. On diverse standard simulation benchmarks including LIBERO~\cite{liu2023libero}, CALVIN~\cite{mees2022calvin}, and Meta-World~\cite{yu2020metaworld}, we compare EDAR-based policies with competitive visuomotor systems. These include strong VQ-based VLA/LLM-style approaches (e.g., FAST, VQ-VLA) as well as continuous-control baselines. Across these benchmarks, EDAR-based policies achieve higher success rates or longer completed task sequences under matched training settings, demonstrating that environment-dependent action representations provide a more effective policy interface than trajectory-centric alternatives. We further validate EDAR on challenging real-world long-horizon manipulation tasks, where it consistently improves over the baseline system under the same training budget. These results establish the empirical advantage of grounding action representations in both executable control structure and environment-conditioned visual consequences.

In summary, our contributions are as follows:
\begin{itemize}
    \item We formulate robotic action representation from an environment-dependent perspective, where action representations explicitly capture both control structure and scene-conditioned visual effects.

    \item We propose \textbf{EDAR} (\textbf{E}nvironment-\textbf{D}ependent \textbf{A}ction \textbf{R}epresentation), a latent action representation framework that combines action reconstruction with visual consequence modeling in a shared image-action token space.

    \item We show that EDAR improves downstream policy learning on diverse simulated and real-robot manipulation benchmarks, with especially strong gains in challenging long-horizon manipulation.
\end{itemize}

\section{Related Work}
\label{sec:rel}

\noindent \textbf{Vision-Language-Action Models.}
Vision-Language-Action (VLA) models provide a powerful paradigm for grounding visual observations and language instructions into robot actions. 
RT-style systems demonstrated that large sequence models and vision-language priors can be adapted to robotic control~\cite{brohan2022rt1,zitkovich2023rt2}, and subsequent models have advanced this direction through larger heterogeneous datasets, stronger multimodal backbones, efficient adaptation, and improved action decoders~\cite{team2024octo,kim2024openvla,black2024pi0,black2025pi05,shukor2025smolvla,wen2024tinyvla,liu2024robomamba,geminirobotics2025}. 
Despite these advances, the action representation remains a central interface in VLA policies: it defines what the model predicts, how action sequences are structured, and how demonstrations are converted into learnable supervision. 
Thus, beyond scaling data and backbones, designing better action representations is an important path toward more effective robotic policies. 
This motivates action representation learning, which aims to replace raw control trajectories with more structured policy interfaces.

\noindent \textbf{Action Representation Learning.}
Prior work has explored various ways to convert raw control trajectories into more learnable action interfaces. 
Discrete or tokenized approaches represent actions with frequency-space tokens, spline or control-point tokens, vector-quantized codes, latent skills, or hierarchical token spaces for long-horizon behavior~\cite{pertsch2025fast,zhou2025beast,lyu2025omnisat,mete2024quest,wang2025vqvla,liu2026oat,gong2025carp,huang2026mint}. 
Continuous approaches instead preserve the structure of action chunks or latent trajectories, enabling temporally coherent prediction and planning in continuous action spaces~\cite{zhao2023act,wang2026latentvla,wu2026colaflow}. 
Although these methods differ in whether the representation is discrete or continuous, they largely share an action-stream-centric view: the representation is learned to preserve, compress, or regularize the structure of the control trajectory itself. 
As a result, the learned action space is mainly organized by command-level properties such as geometry, smoothness, frequency content, reconstruction fidelity, or multiscale decomposition, while the environment-dependent role of the action remains implicit. 
EDAR instead focuses on this missing dimension. 
It uses continuous latent action tokens, but its key contribution is to ground those tokens in the visual consequences induced by actions under the current scene context, allowing environment-conditioned effects to shape the action representation beyond trajectory structure alone.

\noindent \textbf{World Models for Robot Learning.}
Recent robot learning methods increasingly use future-aware supervision, including future observation prediction, video prediction, and latent world modeling~\cite{li2025uva,zhu2025uwm,cen2025worldvla,kim2026cosmospolicy,pai2026dreamzero}. JEPA-style methods perform prediction in latent space to emphasize abstract dynamics over pixel-level details~\cite{assran2023ijepa,assran2025vjepa2,sun2026vlajepa,vujinovic2025actjepa,maes2026leworldmodel}. These methods demonstrate the value of future prediction for policy learning, planning, and action--world modeling. EDAR is related, but addresses a separate question. Existing world-model approaches typically use actions to condition, generate, or reason about future world states; EDAR instead uses predicted visual consequences to organize the latent action space. This distinction is key to our work: future dynamics are not modeled for their own sake, but are used to make action representations environment-dependent.

\section{Method}

\subsection{Overview}

\begin{figure*}[t]
  \centering
  \includegraphics[width=0.92\linewidth]{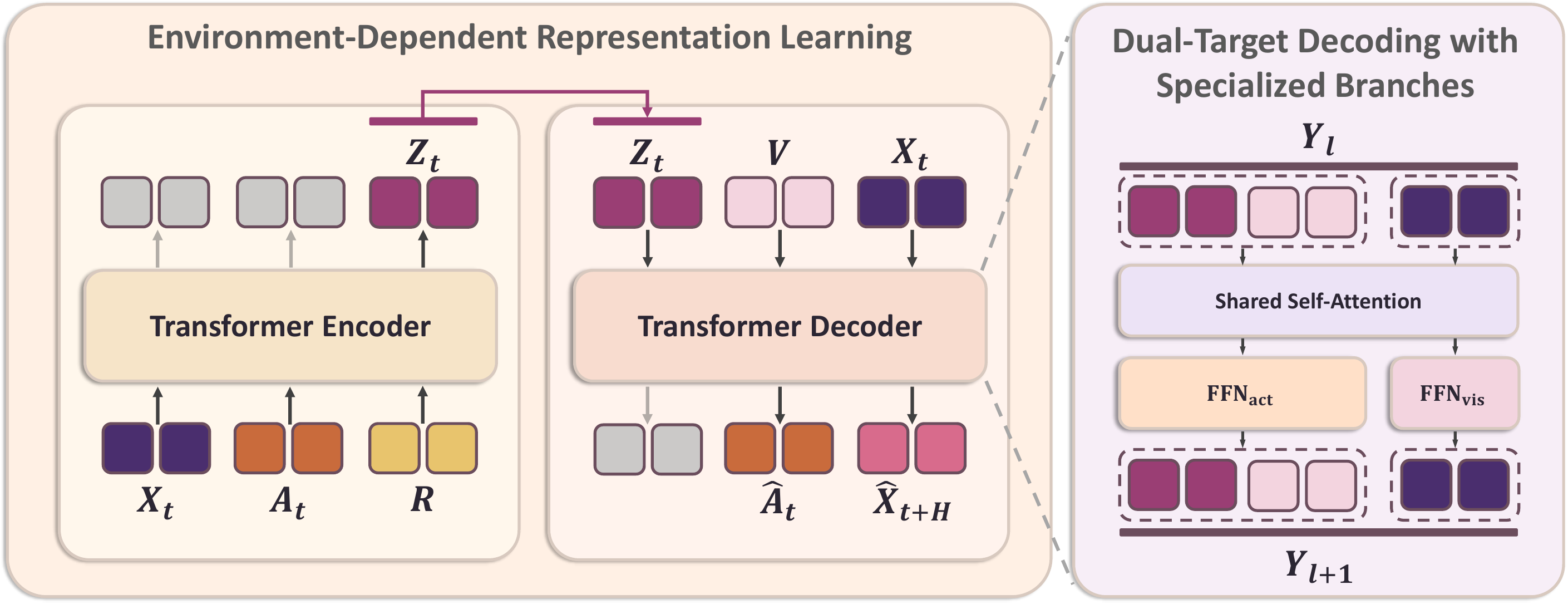}
  \caption{\textbf{Overview of Environment-Dependent Action Representation (EDAR).}
  EDAR encodes the action chunk \(A_t\), current visual tokens \(X_t\), and learnable register tokens \(R\) into continuous latent action tokens \(Z_t\).
  The decoder reconstructs the original action chunk \(\hat{A}_t\) while predicting future visual tokens \(\hat{X}_{t+H}\), forcing the latent action representation to capture both executable control structure and environment-conditioned visual consequences.
  A shared-attention decoder with specialized \(FFN_{\mathrm{act}}\) and \(FFN_{\mathrm{vis}}\) branches handles the heterogeneous action and visual targets.}
  \label{fig:edar_overview}
  \vspace{-3mm}
\end{figure*}

EDAR aims to learn an action representation whose semantics depend on the environment in which the action is executed. 
Given an action chunk 
\(A_t=\{a_{t+h}\}_{h=0}^{H-1}\), current multi-view observations 
\(I_t=\{I_t^{m}\}_{m=1}^{M}\), and future observations after executing the chunk 
\(I_{t+H}=\{I_{t+H}^{m}\}_{m=1}^{M}\), where \(H\) denotes the chunk horizon and \(M\) denotes the number of views, our goal is to encode the action chunk into a fixed number of continuous latent action tokens
\(Z_t=\{z_{t,i}\}_{i=1}^{K}\), where \(K \ll H\). 
These latent tokens serve as the action interface for downstream policy learning.

A useful action representation for manipulation should be both executable and environment-dependent: it must recover the original control trajectory while reflecting the role of the action under the current scene context. 
These requirements expose two potential degeneracies. 
Action-only learning can map similar command trajectories to similar latents despite different environmental effects. 
Simply adding visual context is insufficient: action reconstruction alone does not force the context to shape the action space.

EDAR addresses these issues by making latent action tokens responsible for both execution and effect as shown in Fig.~\ref{fig:edar_overview}. 
Specifically, the current scene participates in latent formation through shared attention between action tokens, visual tokens, and learnable register tokens. 
The resulting latent tokens are then trained with two complementary signals: action reconstruction preserves executability, while visual consequence prediction forces the same tokens to explain how the scene changes after the action is executed. 
In this way, environmental change is not merely provided as input, but directly shapes the learned action space.

\subsection{Environment-Dependent Representation Learning}

We begin by extracting visual tokens from the current multi-view observations using a pretrained image encoder \(E_{\mathrm{im}}\). 
The tokens from all views are concatenated as
\begin{equation}
X_t = [E_{\mathrm{im}}(I_t^{1}); \dots; E_{\mathrm{im}}(I_t^{M})].
\end{equation}
The future visual target is similarly extracted in the same feature space:
\begin{equation}
X_{t+H} = [E_{\mathrm{im}}(I_{t+H}^{1}); \dots; E_{\mathrm{im}}(I_{t+H}^{M})].
\end{equation}
We use visual features rather than pixels as the target, so that the supervision emphasizes task-relevant semantic scene change and interaction dynamics instead of low-level appearance reconstruction.

To form the latent action representation, EDAR jointly encodes the action chunk and the current visual context. 
We instantiate the encoder with a Transformer \(f_{\theta}\) and a set of learnable register tokens 
\(R=\{r_i\}_{i=1}^{K}\). 
The action sequence is first projected into action tokens, which are then concatenated with the register tokens and the current visual tokens. 
The output states corresponding to the register tokens define the latent action tokens:
\begin{equation}
Z_t = f_{\theta}([A_t; R; X_t]).
\end{equation}
The register tokens provide a fixed-size interface for variable-length action chunks, while shared self-attention allows action content and scene context to interact before the latent representation is formed. 
This design lets the latent action tokens be formed with respect to the current environment, rather than being determined solely by the structure of the action chunk.

However, context-conditioned latent formation alone is not sufficient. 
If the only objective is to reconstruct \(A_t\), the representation can still be optimized around trajectory geometry, temporal variation, or other command-level properties. 
To make environment dependence intrinsic to the latent space, EDAR requires \(Z_t\) to explain not only the executed command, but also its consequence under the current scene context. 
A decoder \(g_{\phi}\) takes \(Z_t\), \(X_t\), and a learnable query embedding \(V\) as input, and predicts both the original action chunk and the future visual tokens:
\begin{equation}
(\hat{A}_t, \hat{X}_{t+H}) = g_{\phi}(Z_t, V, X_t).
\end{equation}
Here, \(V\) serves as a learnable action query that helps decode the latent tokens into the executable action chunk.

The action reconstruction loss preserves the executable structure of the latent representation:
\begin{equation}
\mathcal{L}_{\mathrm{act}} =
\frac{1}{H}\sum_{h=0}^{H-1}
\left\|a_{t+h} - \hat{a}_{t+h}\right\|_2^2.
\end{equation}
The visual consequence loss encourages the latent to capture the environment-conditioned effect of the action:
\begin{equation}
\mathcal{L}_{\mathrm{eff}} =
\frac{1}{M}\sum_{m=1}^{M}
\left(
1 - \cos\bigl(
\operatorname{sg}(X_{t+H}^{m}),
\hat{X}_{t+H}^{m}
\bigr)
\right),
\end{equation}
where \(M\) is the number of multi-view observations, \(\operatorname{sg}(\cdot)\) denotes stop-gradient, and \(\cos(\cdot,\cdot)\) denotes cosine similarity. 
The stop-gradient operation keeps the visual target space fixed, so that the action representation is shaped by pretrained visual semantics.

The full representation learning objective is
\begin{equation}
\mathcal{L}_{\mathrm{EDAR}} =
\mathcal{L}_{\mathrm{act}} + \lambda \mathcal{L}_{\mathrm{eff}}.
\end{equation}
The two terms remove different degeneracies. 
Without \(\mathcal{L}_{\mathrm{act}}\), the latent may become a representation of scene change without remaining a reliable action interface. 
Without \(\mathcal{L}_{\mathrm{eff}}\), the latent may collapse back to an action-only trajectory code. 
Their combination makes \(Z_t\) both executable and environment-dependent.

\subsection{Dual-Target Decoding with Specialized Branches}

The two learning signals in EDAR correspond to heterogeneous output spaces: action reconstruction is performed in continuous control space, while visual consequence prediction is performed in pretrained visual feature space. 
A naive decoder that applies the same token-wise transformation to both targets can introduce optimization conflict, since the two targets require different output structures and levels of abstraction. 
At the same time, using two separate decoders would weaken the coupling between execution and effect, reducing the ability of visual consequence prediction to shape the same latent action space used for control.

To balance these requirements, EDAR uses a decoder with shared attention and token-specialized feed-forward branches. 
The latent action tokens \(Z_t\) and current visual tokens \(X_t\) are first projected into a common hidden space,
\begin{equation}
Y^{0} = [\Pi_z(Z_t); \Pi_x(X_t)].
\end{equation}
At each decoder layer, tokens interact through a shared self-attention module with an asymmetric causal mask, while the feed-forward update is specialized according to token type:
\begin{equation}
\begin{aligned}
\tilde{Y}^{\ell} &= \mathrm{Attn}^{\ell}(Y^{\ell}), \\
Y^{\ell+1} &=
\left[
\mathrm{FFN}^{\ell}_{\mathrm{act}}(\tilde{Y}^{\ell}_{Z});
\mathrm{FFN}^{\ell}_{\mathrm{vis}}(\tilde{Y}^{\ell}_{X})
\right].
\end{aligned}
\end{equation}
Here, \(\tilde{Y}^{\ell}_{Z}\) and \(\tilde{Y}^{\ell}_{X}\) denote the hidden states at latent-action-token and visual-token positions. 
After the final layer, separate output heads predict the two targets:
\begin{equation}
\hat{A}_t = o_{\mathrm{act}}(Y_Z^{L}), 
\qquad
\hat{X}_{t+H} = o_{\mathrm{vis}}(Y_X^{L}).
\end{equation}

This design keeps the two targets coupled through shared attention, while allowing each target to use a specialized feed-forward transformation. 
As a result, the decoder supports both control fidelity and consequence modeling without forcing a single output head to serve incompatible prediction spaces. 
Importantly, the future prediction branch is not used to build a pixel-level world model or to perform rollout-based planning. 
Its role is to provide an effect-sensitive learning signal that organizes the latent action space according to environment-conditioned consequences.

\subsection{Downstream Policy Evaluation}

After representation learning, EDAR can be used as an action interface for downstream policies. 
Instead of predicting raw action chunks, the policy predicts latent action tokens \(Z_t\), which are then decoded into executable controls by the learned action decoder. 
This evaluation tests whether the environment-dependent latent space is easier to model than the raw action space or an action-only latent space.

We instantiate the downstream policy with a flow-matching policy in the EDAR latent space. 
Given visual observations and language instruction, the policy receives a visuolinguistic condition \(\mathbf{c}\). 
Let \(Z_1 = Z_t\) denote the target latent action tokens obtained from EDAR, and sample a Gaussian noise endpoint \(Z_0 \sim \mathcal{N}(0, I)\). 
For interpolation time \(\tau \in [0,1]\), define:
\begin{equation}
Z_{\tau} = (1-\tau)Z_0 + \tau Z_1.
\end{equation}
The policy learns a velocity field \(v_{\psi}(Z_{\tau}, \tau, \mathbf{c})\) in latent space by minimizing:
\begin{equation}
\mathcal{L}_{\mathrm{FM}}
=
\mathbb{E}_{\tau,\, Z_0,\, Z_1}
\Big[
\big\|
v_{\psi}(Z_{\tau}, \tau, \mathbf{c}) - (Z_1 - Z_0)
\big\|_2^2
\Big].
\end{equation}
At inference, the policy denoises Gaussian noise into latent action tokens via the learned velocity field.
The EDAR decoder then maps the generated latent tokens back to executable robot controls. 
The downstream policy is not the main contribution of EDAR; rather, it serves as an evaluation protocol for whether the learned action representation provides a more effective policy interface.

\section{Experiments}

We design our experiments to test the central hypothesis of EDAR: action representations for robotic manipulation should be environment-dependent. Specifically, we examine whether environment-dependent action representations provide a more effective policy interface than trajectory-centric alternatives.

We organize the experiments around three questions:
\begin{itemize}
    \item \textbf{Environment-dependent advantage.}
    Does EDAR outperform trajectory-centric action representations under matched training settings?

    \item \textbf{Robustness across settings.}
    Are the gains consistent across different environments, task distributions, and policy backbones?

    \item \textbf{Long-horizon and real-world stability.}
    Does EDAR improve execution stability in long-horizon and real-world manipulation tasks under the same downstream policy-training budget?
\end{itemize}

\subsection{Simulation Experiments}
\label{subsec:simulation}

We evaluate EDAR in simulation on three widely used robotic manipulation benchmarks: LIBERO~\citep{liu2023libero}, CALVIN~\citep{mees2022calvin}, and Meta-World~\citep{yu2020metaworld}. 
LIBERO evaluates language-conditioned manipulation across multiple task suites with different forms of generalization, while CALVIN focuses on long-horizon language-conditioned control in a changing tabletop environment, and Meta-World focuses on multi-task manipulation across tasks of varying difficulty.
Together, these benchmarks allow us to test both the controlled advantage of environment-dependent action representation and its robustness across different task structures.

\subsubsection{Evaluation on the LIBERO Benchmark}
\label{subsubsec:libero}

\begin{table*}[t]
\centering
\caption{
\textbf{Comparison on LIBERO.}
We report success rates (\%) on the four LIBERO task suites.
Methods are organized by their role in the evaluation: representative VLA systems are provided for context, while controlled comparisons isolate the effect of replacing the action interface with EDAR under the same policy backbone.
Avg. denotes the average success rate.
}
\label{tab:libero}
\small
\setlength{\tabcolsep}{4.5pt}
\renewcommand{\arraystretch}{0.88}
\begin{tabular}{lccccc}
\toprule
Method & Spatial & Object & Goal & Long & Avg. \\
\midrule

Octo~\citep{team2024octo} & 78.9 & 85.7 & 84.6 & 51.1 & 75.1 \\
OpenVLA~\citep{kim2024openvla} & 84.7 & 88.4 & 79.2 & 53.7 & 76.5 \\
VQ-VLA~\citep{wang2025vqvla} & -- & -- & 75.2 & 60.0 & -- \\
Diffusion Policy~\citep{chi2023diffusion} & 78.3 & 92.5 & 68.3 & 50.5 & 72.4 \\
VLA-4D~\citep{zhang20254d} & 93.8 & 92.8 & 95.6 & 86.5 & 92.2 \\
CogACT~\citep{li2024cogact} & 97.2 & 98.0 & 90.2 & 88.8 & 93.6 \\
OmniSAT~\citep{lyu2025omnisat} & 94.1 & 98.7 & 94.6 & 86.0 & 93.4 \\

\midrule
SmolVLA-FAST & 87.0 & 93.0 & 90.0 & 68.0 & 84.5 \\
SmolVLA-VQ-VLA & 90.0 & 94.0 & 86.0 & 71.0 & 85.3 \\
SmolVLA~\citep{shukor2025smolvla} & 90.0 & 96.0 & 92.0 & 71.0 & 87.3 \\
SmolVLA-VAE & 90.8 & 97.6 & 92.8 & 75.0 & 89.1 \\
\textbf{SmolVLA-EDAR (Ours)} & \textbf{94.6} & \textbf{98.0} & \textbf{93.8} & \textbf{80.6} & \textbf{91.8} \\

\midrule
$\pi_0$-FAST~\citep{pertsch2025fast} & 96.4 & 96.8 & 88.6 & 60.2 & 85.5 \\
$\pi_0$~\citep{black2024pi0} & 96.8 & 98.8 & 95.8 & 85.2 & 94.2 \\
\textbf{$\pi_0$-EDAR (Ours)} & \textbf{98.2} & \textbf{99.2} & \textbf{96.8} & \textbf{94.2} & \textbf{97.1} \\

\bottomrule
\end{tabular}
\end{table*}

\paragraph{Setup.}
We first evaluate EDAR on LIBERO~\citep{liu2023libero}, which includes four manipulation suites testing generalization over spatial relations, object categories, task goals, and long-horizon behaviors.

Our controlled study uses SmolVLA~\citep{shukor2025smolvla} as a common backbone, keeping the data, preprocessing, and training protocol fixed while varying only the action interface. We compare the original SmolVLA action head with trajectory-tokenization alternatives, including SmolVLA-FAST and SmolVLA-VQ-VLA, and our environment-dependent continuous latent action tokens. To isolate the role of environment dependence, we further include SmolVLA-VAE, which keeps the learned latent-action interface but removes environment information and visual consequence supervision. We also integrate EDAR into the stronger diffusion-based $\pi_0$ model~\citep{black2024pi0}. Representative VLA systems are reported as contextual baselines, since they differ in backbone, pretraining data, action heads, and training recipes.

\paragraph{Results.}
Table~\ref{tab:libero} reports the LIBERO results. Under the controlled SmolVLA setting, EDAR improves the average success rate from 87.3\% to 91.8\% (+4.5) without changing the backbone, data, or training protocol, with the largest gain on LIBERO-Long from 71.0\% to 80.6\% (+9.6).

It also outperforms SmolVLA-FAST and SmolVLA-VQ-VLA across all four suites, providing a system-level comparison against representative action-interface alternatives under the same VLA backbone and demonstrating the effectiveness of an environment-dependent continuous latent interface for downstream policy learning.

The comparison with SmolVLA-VAE further shows that while a learned latent-action interface is already beneficial, explicitly conditioning it on the environment and supervising it with visual consequences provides additional benefits, improving the average success rate from 89.1\% to 91.8\% and LIBERO-Long from 75.0\% to 80.6\%.

EDAR also improves the stronger $\pi_0$ backbone, increasing the average success rate from 94.2\% to 97.1\%, with the largest gain on LIBERO-Long from 85.2\% to 94.2\% (+9.0), suggesting that EDAR can serve as a plug-in action representation component across different VLA backbones.

Overall, these results support the central claim that effective manipulation action interfaces should preserve executable control structure while being grounded in environment-conditioned consequences.

\subsubsection{Evaluation on the CALVIN ABCD$\rightarrow$D Benchmark}
\label{subsubsec:calvin}

\begin{table*}[t]
\centering
\caption{
\textbf{Comparison on CALVIN ABCD$\rightarrow$D.}
We report the success rate (\%) of completing at least $k$ consecutive instructions over 1000 five-instruction chains, and the average completed sequence length.
The upper block lists representative baselines for context; for these rows, Avg. Len. is the scalar reported by the original paper and is not recomputed from rounded per-step percentages.
The lower block reports our controlled comparisons, where each baseline and EDAR variant share the same training data, schedule, and evaluation protocol. $*$ denotes models with pretraining.
}
\label{tab:calvin}
\small
\setlength{\tabcolsep}{4.0pt}
\renewcommand{\arraystretch}{0.88}
\begin{tabular}{l ccccc c}
\toprule
Method & 1/5 & 2/5 & 3/5 & 4/5 & 5/5 & Avg. Len. \\
\midrule
Diff-P-CNN~\citep{chi2023diffusion} & 86.3 & 72.7 & 60.1 & 51.2 & 41.7 & 3.16 \\
RoboFlamingo~\citep{li2023roboflamingo} & 96.4 & 89.6 & 82.4 & 74.0 & 66.0 & 4.09 \\
DeeR-VLA~\citep{yue2024deervla} & 99.1 & 93.3 & 82.1 & 74.6 & 63.8 & 4.13 \\
GR-1~\citep{wu2023gr1} & 94.9 & 89.6 & 84.4 & 78.9 & 73.1 & 4.21 \\
MDT~\citep{reuss2024mdt} & 98.6 & 95.8 & 91.6 & 86.2 & 80.1 & 4.52 \\
MoDE~\citep{reuss2024mode} & 97.1 & 92.5 & 87.9 & 83.5 & 77.9 & 4.39 \\
MINT-4B~\citep{huang2026mint} & 97.4 & 94.2 & 91.7 & 88.2 & 86.1 & 4.57 \\
FLOWER-VLA~\citep{reuss2025flower} & 98.9 & 96.7 & 93.9 & 90.2 & 85.5 & 4.62 \\
FLOWER-VLA$^*$~\citep{reuss2025flower} & 99.2 & 96.9 & 96.9 & 92.3 & 88.3 & 4.67 \\
\midrule
SmolVLA~\citep{shukor2025smolvla} & 92.5 & 78.9 & 65.1 & 53.2 & 44.7 & 3.34 \\
\textbf{SmolVLA-EDAR (Ours)} & \textbf{94.4} & \textbf{84.2} & \textbf{75.3} & \textbf{67.0} & \textbf{58.9} & \textbf{3.80} \\
\midrule
FLOWER-VLA (FSDP repro.) & 98.0 & 95.1 & 91.7 & 88.2 & 84.7 & 4.58 \\
\textbf{FLOWER-VLA-EDAR (Ours)} & \textbf{99.7} & \textbf{98.2} & \textbf{96.0} & \textbf{92.5} & \textbf{88.1} & \textbf{4.75} \\
\bottomrule
\end{tabular}
\end{table*}

\paragraph{Setup.}
We further evaluate EDAR on CALVIN ABCD$\rightarrow$D~\citep{mees2022calvin}, a long-horizon language-conditioned manipulation benchmark.
Each episode contains a chain of five instructions executed without resetting the environment, so later subtasks depend on the scene state produced by earlier actions.
This makes CALVIN a suitable testbed for evaluating whether an action representation remains reliable across compounding scene transitions.

We integrate EDAR into two policy backbones: SmolVLA~\citep{shukor2025smolvla} and FLOWER-VLA~\citep{reuss2025flower}. In both cases, EDAR replaces the original action interface with environment-dependent latent action tokens, while the remaining policy architecture and training protocol are kept matched to the corresponding baseline.
For FLOWER-VLA, we report both the original published results as external context and our FSDP reproduction as the controlled baseline for comparison with FLOWER-VLA-EDAR.

\paragraph{Results.}
Table~\ref{tab:calvin} reports the CALVIN results. In the matched SmolVLA setting, EDAR improves the average completed sequence length from 3.34 to 3.80. The improvement becomes larger as the instruction chain progresses, with gains of +1.9, +5.3, +10.2, +13.8, and +14.2 points from 1/5 to 5/5 success. This indicates that EDAR is especially beneficial when the policy must remain coherent over multiple environment transitions.

EDAR also improves the stronger FLOWER-VLA backbone under the controlled FSDP reproduction, increasing the average completed sequence length from 4.58 to 4.75. The gains are consistent across all chain positions, from +1.7 at 1/5 to +3.4 at 5/5. Since the original published FLOWER-VLA numbers come from a different distributed-training pipeline, we treat them as external context rather than the primary controlled comparison.

Overall, CALVIN confirms the long-horizon benefit of environment-dependent action representation. By grounding latent action tokens in expected scene changes, EDAR improves not only local instruction execution but also robustness over extended action chains where earlier actions determine the states faced by later ones.

\begin{table*}[t]
\centering
\caption{\textbf{Comparison on Meta-World.} We report success rates (\%) across tasks grouped by difficulty. Avg. denotes the unweighted mean over the four difficulty groups.}
\label{tab:metaworld}
\small
\setlength{\tabcolsep}{4.5pt}
\renewcommand{\arraystretch}{0.88}
\begin{tabular}{lccccc}
\toprule
Method & Easy & Medium & Hard & Very Hard & Avg. \\
\midrule

Diffusion Policy~\citep{chi2023diffusion} 
& 23.1 & 10.7 & 1.9 & 6.1 & 10.5 \\

TinyVLA-H~\citep{wen2024tinyvla} 
& 77.6 & 21.5 & 11.4 & 15.8 & 31.6 \\

$\pi_0$~\citep{black2024pi0} 
& 80.4 & 40.9 & 36.7 & 44.0 & 50.5 \\

RoboTron Mani~\citep{yan2024robotron}
& 85.5 & 67.7 & 76.7 & 87.0 & 77.7 \\

\midrule

SmolVLA(0.45B, our repro.) 
& 88.6 & 71.8 & 55.0 & 62.0 & 69.4 \\

\textbf{SmolVLA-EDAR (Ours)} 
& \textbf{89.3} & \textbf{76.4} & \textbf{78.3} & \textbf{78.0} & \textbf{80.5} \\

\midrule

Evo-1~\citep{lin2025evo}
& 89.2 & 76.8 & 77.2 & 79.2 & 80.6 \\

\textbf{Evo-1-EDAR (Ours)} 
& \textbf{93.2} & \textbf{91.8} & \textbf{83.3} & \textbf{88.0} & \textbf{89.1} \\

\bottomrule
\end{tabular}
\end{table*}

\subsubsection{Evaluation on the Meta-World Benchmark}

\paragraph{Setup.}
We further evaluate EDAR on Meta-World~\citep{yu2020metaworld}, a multi-task robotic manipulation benchmark. Following prior work, we group tasks into four difficulty levels: Easy, Medium, Hard, and Very Hard, and report the unweighted average over the four groups.

We integrate EDAR into two policy backbones, SmolVLA and Evo-1~\citep{lin2025evo}, for evaluation on Meta-World. For both backbones, EDAR replaces the original continuous action interface with environment-dependent latent action tokens, while the dataset, data processing pipeline, training schedule, evaluation protocol, and checkpoint selection are kept matched to their respective reproduced baselines. This controlled setup allows us to isolate the effect of the proposed action representation across different policy backbones.

We also include results from prior work, including Diffusion Policy~\citep{chi2023diffusion}, TinyVLA~\citep{wen2024tinyvla}, and $\pi_0$~\citep{black2024pi0}, as external context. These numbers may differ in model scale, training data, preprocessing, and evaluation details, and are not intended as strictly controlled comparisons.

\paragraph{Results.}
Table~\ref{tab:metaworld} reports the Meta-World results.
Under the controlled SmolVLA setting, EDAR improves the average score from 69.4 to 80.5, yielding a consistent absolute gain of +11.1 points.
The improvement is modest on Easy tasks (+0.7), but becomes substantially larger on Medium (+4.6), Hard (+23.3), and Very Hard (+16.0) tasks. This shows that replacing the original SmolVLA action interface with EDAR provides clear gains, especially in more challenging task groups.

On the stronger Evo-1~\citep{lin2025evo} backbone, EDAR delivers a substantial further improvement, increasing the average score from 80.6 to 89.1, corresponding to an absolute gain of +8.5 points and bringing its performance into the state-of-the-art range on this benchmark. This demonstrates that the effectiveness of EDAR is not specific to SmolVLA and generalizes to a stronger policy backbone.

Compared with external baselines from prior work, Evo-1-EDAR achieves the highest average performance in this table.
These results indicate that EDAR remains effective on Meta-World, a broad multi-task manipulation benchmark with diverse task difficulties.

\begin{figure*}[t]
\centering
\includegraphics[width=0.9\linewidth]{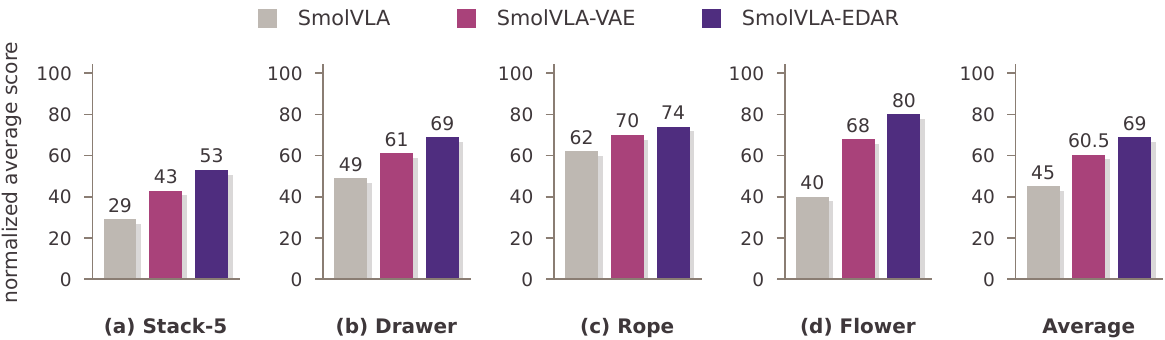}
\caption{
\textbf{Real-world long-horizon manipulation results.}
We report normalized average score on four real-world tasks: Stack-5, Drawer, Rope, and Flower, together with the average score across tasks.
}
\label{fig:real_world_results}
\vspace{-5mm}
\end{figure*}

\subsection{Long-Horizon Real-World Manipulation}
\label{subsec:real_world}

\begin{figure}[t]
\centering

\begin{subfigure}[t]{1.0\columnwidth}
    \centering
    \includegraphics[width=\linewidth]{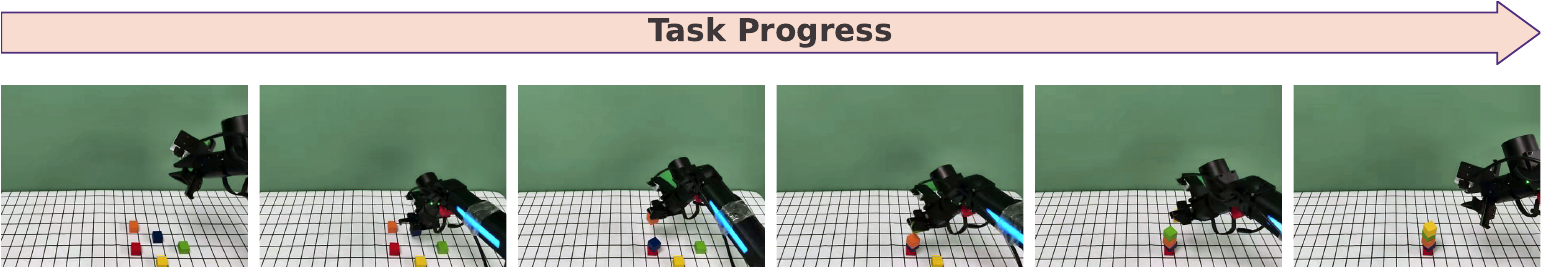}
    \caption{Stack-5: stack blocks in yellow-green-orange-blue-red order.}
    \label{fig:real_world_stack5}
\end{subfigure}


\begin{subfigure}[t]{1.0\columnwidth}
    \centering
    \includegraphics[width=\linewidth]{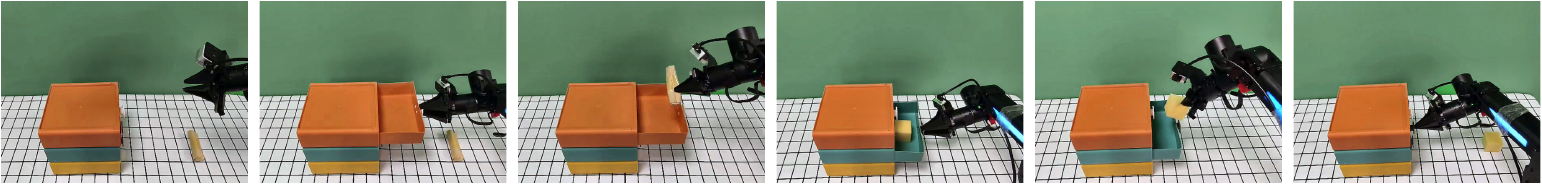}
    \caption{Drawer: open the top drawer, put the tissue into the top drawer and close it; open the middle drawer, take the sponge out and close the drawer.}
    \label{fig:real_world_drawer}
\end{subfigure}


\begin{subfigure}[t]{1.0\columnwidth}
    \centering
    \includegraphics[width=\linewidth]{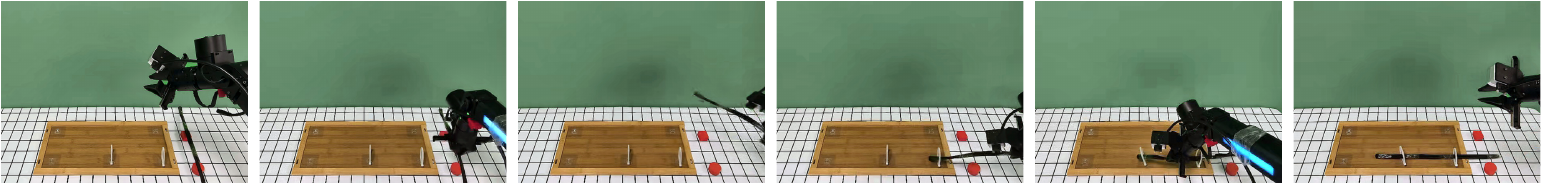}
    \caption{Rope: pick the rope, thread it into the first ring, drop it, pick it from the other side of the ring, and thread it into the second ring.}
    \label{fig:real_world_rope}
\end{subfigure}


\begin{subfigure}[t]{1.0\columnwidth}
    \centering
    \includegraphics[width=\linewidth]{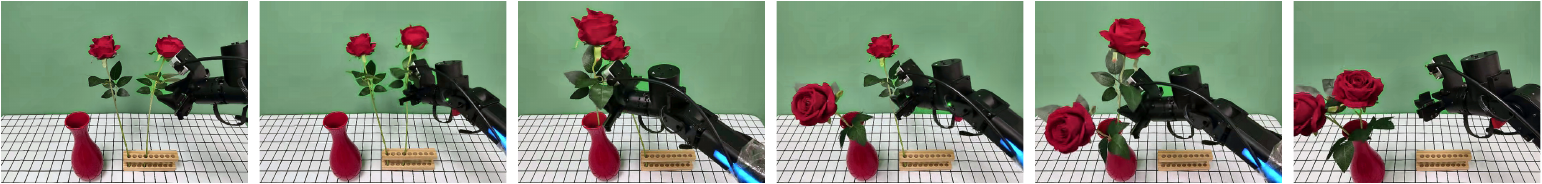}
    \caption{Flower: pick one flower and insert it into the vase; pick the remaining flower and insert it into the vase.}
    \label{fig:real_world_flower}
\end{subfigure}
\vspace{-2mm}
\caption{
\textbf{Qualitative results of long-horizon real-world tasks.}
Key frames for the four long-horizon tasks. Each subfigure shows the progression from start to completion.
}
\label{fig:real_world_task_progress}

\end{figure}

\paragraph{Setup.}
We evaluate EDAR on four long-horizon real-world manipulation tasks:
\textbf{Stack-5} (stack four cubes sequentially on a red base cube to form a five-cube tower), 
\textbf{Drawer} (open the top drawer, place a tissue inside, and close it; then open the middle drawer and remove a sponge and close the drawer), 
\textbf{Rope} (thread a rope through two rings sequentially), and 
\textbf{Flower} (pick and place two roses into a vase, one at a time).
These tasks require sequential completion of multiple manipulation stages, where later stages strongly depend on the intermediate scene state produced by earlier actions, as shown in Figure~\ref{fig:real_world_task_progress}. The experiments are conducted with a single ARX R5 arm equipped with one third-person camera and one wrist camera. For each task, we collect 100 teleoperated demonstrations and train a task-specific policy.

As the baseline, SmolVLA is trained independently on each task.
For the latent-action variants, we compare SmolVLA-VAE, an environment-independent VAE-style action representation, and SmolVLA-EDAR, which uses action reconstruction and visual consequence grounding.
Both representations are first pretrained on Open X-Embodiment~\cite{o2024open} following the VQ-VLA~\cite{wang2025vqvla} setup, then fine-tuned on the target task demonstrations and integrated into SmolVLA.
All policy variants are trained for the same number of task-specific steps, isolating the effect of the action representation.

For evaluation, each task is decomposed into 4--8 sub-steps according to its execution horizon.
Each completed sub-step contributes one point, and the total score is normalized to a 0--100 scale, where 100 indicates that all sub-steps are completed. For each task, we conduct 10 evaluation runs and report the average normalized score across runs.

\paragraph{Results and analysis.}
Fig.~\ref{fig:real_world_results} reports the normalized task scores on the four real-world tasks. 
SmolVLA-EDAR achieves the best performance across all tasks, improving the average score from 45.0\% for SmolVLA and 60.5\% for SmolVLA-VAE to 69.0\%.
Compared with SmolVLA, EDAR improves Stack-5, Drawer, Rope, and Flower by +24.0, +20.0, +12.0, and +40.0 points, respectively.

EDAR further improves over SmolVLA-VAE by +8.5 points on average, showing that the gain is not only due to learning a structured latent action representation.
By grounding action latents in current visual context and future visual consequences, EDAR better handles changing scene states and reduces compounding errors during long-horizon execution. The consistent improvements across stacking, articulated-object manipulation, and deformable-object manipulation suggest that EDAR is particularly beneficial in real-world sequential manipulation scenarios where intermediate progress directly affects the final outcome.

Representative qualitative examples of successful real-world executions are shown in Fig.~\ref{fig:real_world_task_progress} for reference.

\subsection{Ablation Studies}
\label{subsec:ablations}

We analyze the design choices of EDAR in Table~\ref{tab:edar_ablation}.
Unless otherwise specified, all variants use SmolVLA as the downstream policy backbone and are evaluated on LIBERO under the same data, training schedule, and evaluation protocol as the main experiments.

\begin{table*}[t]
\centering
\scriptsize
\setlength{\tabcolsep}{2.2pt}
\renewcommand{\arraystretch}{1.05}

\begin{minipage}[t]{0.315\textwidth}
\centering
\begin{tabular*}{\linewidth}{@{\extracolsep{\fill}}lcc@{}}
\toprule
Action encoding & Avg. $\uparrow$ & $\Delta$ \\
\midrule
Full EDAR
& \cellcolor{gray!18}\textbf{91.8} & 0.0 \\
w/o Enc. env. ctx.
& 90.9 & -0.9 \\
\bottomrule
\end{tabular*}

\raggedright
\footnotesize
\textbf{(a) Env.-conditioned encoding.}
Ablates current environment context in action encoding.
\end{minipage}
\hfill
\begin{minipage}[t]{0.315\textwidth}
\centering
\begin{tabular*}{\linewidth}{@{\extracolsep{\fill}}lcc@{}}
\toprule
Effect supervision & Avg. $\uparrow$ & $\Delta$ \\
\midrule
Full EDAR
& \cellcolor{gray!18}\textbf{91.8} & 0.0 \\
w/o future pred.
& 90.1 & -1.7 \\
\bottomrule
\end{tabular*}

\raggedright
\footnotesize
\textbf{(b) Consequence prediction.}
Ablates the future visual prediction objective.
\end{minipage}
\hfill
\begin{minipage}[t]{0.315\textwidth}
\centering
\begin{tabular*}{\linewidth}{@{\extracolsep{\fill}}lcc@{}}
\toprule
Decoder design & Avg. $\uparrow$ & $\Delta$ \\
\midrule
Specialized branches
& \cellcolor{gray!18}\textbf{91.8} & 0.0 \\
Single decoder
& 90.2 & -1.6 \\
Two decoders
& 88.7 & -3.1 \\
\bottomrule
\end{tabular*}

\raggedright
\footnotesize
\textbf{(c) Decoder design.} Compares decoder designs for joint action-effect prediction.
\end{minipage}

\caption{
\textbf{Ablation study of EDAR on LIBERO.}
Each panel isolates one design question using Full EDAR as the reference setting:
environment-conditioned action encoding, future consequence prediction, and dual-target decoder design.
Gray cells denote the reference setting in each panel, and $\Delta$ is computed relative to that setting.
}
\label{tab:edar_ablation}
\end{table*}

\paragraph{Environment-conditioned action encoding.}
Tab.~\ref{tab:edar_ablation}a evaluates the effect of current environment context in action encoding.
Removing encoder-side environment context reduces the average success rate from 91.8\% to 90.9\%.
This drop directly supports our motivation that action semantics in manipulation are environment-dependent.
The same motor trajectory can induce different effects under different scene states; therefore, latent action tokens should be formed with access to the current environment rather than learned as scene-agnostic trajectory embeddings.

\paragraph{Consequence prediction supervision.}
Tab.~\ref{tab:edar_ablation}b evaluates the effect of future visual prediction as auxiliary supervision.
Removing this objective decreases the average success rate from 91.8\% to 90.1\%, even though the action encoder still uses current environment context.
This indicates that environment-conditioned encoding alone is insufficient to make the latent action space effect-aware.
By predicting future visual consequences, EDAR encourages latent action tokens to preserve information about action-induced scene changes, such as object motion, contact outcomes, and task-relevant transitions.

\paragraph{Dual-target decoder design.}
Tab.~\ref{tab:edar_ablation}c compares decoder designs for joint action-effect prediction.
EDAR uses specialized branches, where action and visual tokens interact through shared attention while target-specific FFNs handle their different output spaces.
Replacing this design with a single shared decoder reduces performance from 91.8\% to 90.2\%, suggesting interference between continuous action reconstruction and visual feature prediction.
Using two fully separate decoders further drops performance to 88.7\%, indicating that decoupling the two targets weakens action--effect interaction.
These results show that future consequence prediction is most effective when action and effect representations are coupled through shared attention but decoded with target-specific transformations.

\begin{figure}[t]
\centering
\includegraphics[width=0.95\columnwidth]{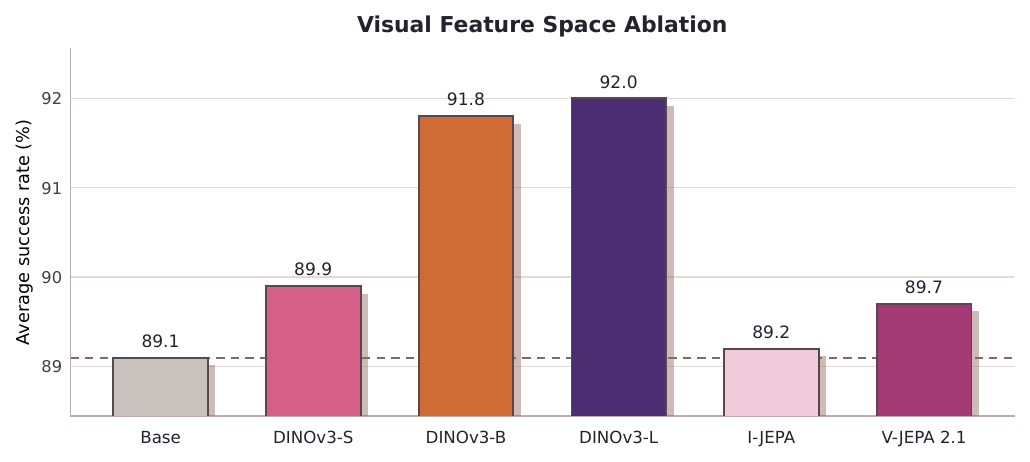}
\caption{
\textbf{Ablation of visual feature spaces for future consequence prediction.}
All variants use the same EDAR architecture and differ only in the visual encoder used to produce future feature targets. We report the average success rate on LIBERO.
}
\label{fig:image_encoder_ablation}
\end{figure}

\paragraph{Visual feature encoder.}
Fig.~\ref{fig:image_encoder_ablation} studies the choice of pretrained visual encoder used to construct future feature targets.
All variants use the same EDAR architecture, downstream SmolVLA policy, and training schedule, and differ only in the feature space used for future consequence prediction.
We first observe a clear scaling trend within the DINOv3 family~\cite{simeoni2025dinov3}.
Compared with the action-only baseline at 89.1\%, DINOv3-S improves the average success rate to 89.9\%, DINOv3-B further increases it to 91.8\%, and DINOv3-L reaches 92.0\%.
This suggests that stronger image feature targets provide more informative supervision for shaping the latent action space.

In contrast, JEPA-based targets do not show the same benefit.
I-JEPA achieves 89.2\%, close to the baseline, and V-JEPA 2.1 reaches 89.7\%, still below the DINOv3-B/L variants.
This may be because EDAR relies on endpoint dense-feature prediction, where the target space should provide spatially stable local supervision for action-induced scene changes. As visualized in Appendix~\ref{sec:visual_context_quality}, DINOv3 features show clearer object-background separation than V-JEPA2.1 features, making them more effective visual targets for organizing action latents.

\section{Conclusion}

This work studies action representation learning for robotic manipulation from an environment-dependent perspective.
We argue that the meaning of an action should not be defined by its control trajectory alone, but by the consequence it induces in a specific scene.
Based on this idea, we propose EDAR, which conditions latent action encoding on the current environment and supervises the latent space with future visual consequences.
The resulting representation preserves executability while encouraging action tokens to capture interaction-relevant scene changes.
Across simulation and real-world experiments, EDAR provides a more effective action interface for downstream policy learning than trajectory-only representations.
The ablation studies further show that environment-conditioned encoding, future consequence prediction, and the dual-target decoder design each contribute to the final performance.
These findings support the broader view that action spaces for embodied agents should be grounded in action--environment interaction, rather than only in motor trajectory structure.
\clearpage
{
    \small
    \bibliographystyle{ieeenat_fullname}
    \bibliography{main}
}

\clearpage
\appendix
\clearpage
\maketitlesupplementary

\renewcommand\thesection{\Alph{section}}
\setcounter{section}{0}

\renewcommand{\thefigure}{\Alph{figure}}
\setcounter{figure}{0}
\renewcommand{\thetable}{\Alph{table}}
\setcounter{table}{0}


\section{Implementation Details}
This section provides details on the architecture and training procedures across all benchmarks. We describe the details as follows.
\subsection{Architecture}

\begin{figure*}[t]
    \centering
    \includegraphics[width=0.95\linewidth]{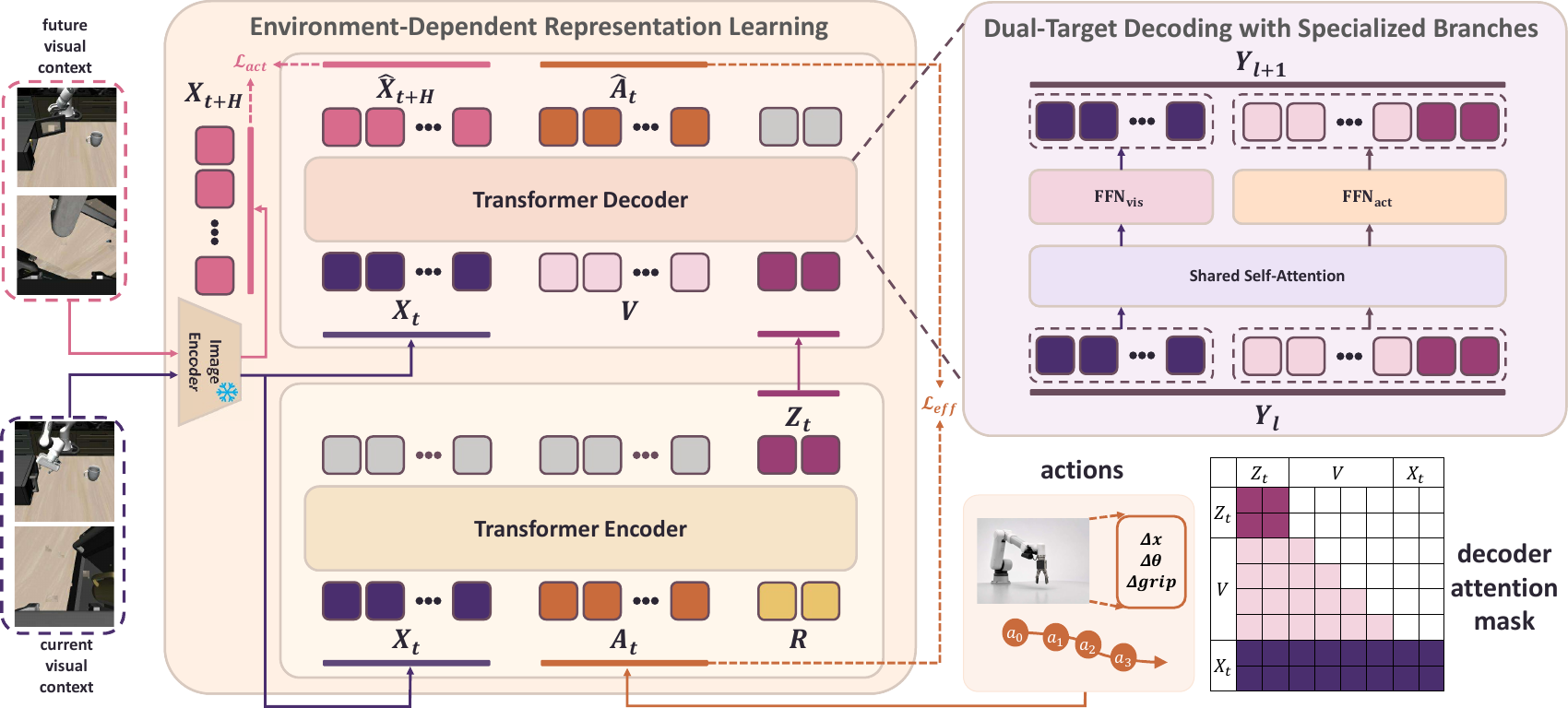}
    \caption{\textbf{Detailed EDAR mechanism.} EDAR forms latent action tokens through full-attention interaction among current visual tokens, action tokens, and learnable register tokens, then decodes them to reconstruct actions and predict future visual features. The bottom-right panels illustrate the decoder asymmetric mask, where visual prediction tokens can attend to the full token set, while action reconstruction remains causal and independent of visual prediction tokens.}
    \label{fig:edar_detailed_mechanism}
\end{figure*}

\paragraph{Overview.}
EDAR is implemented as an 8-layer Transformer encoder paired with an 8-layer Transformer decoder, each with a hidden size of 512, 8 attention heads, and feed-forward networks with a 4$\times$ expansion ratio. We use RMSNorm~\cite{zhang2019root} as the normalization layer for all Transformer blocks. The image encoder is instantiated as DINOv3-Base and is used to extract visual features from image observations, as shown in Fig.~\ref{fig:edar_detailed_mechanism}. The dimensionality of the action latent is set equal to the original action dimension.

\paragraph{Visual feature encoder and feature space.}
EDAR uses the current multi-view observation as visual context and predicts both the action chunk and a future visual-context target. Standard EDAR uses DINOv3-Base patch features; the DINO encoder is used as a frozen feature extractor in the action representation implementation. The decoder predicts actions and future DINOv3 patch features. The image prediction loss is implemented as a cosine-distance loss between predicted and target DINO features, scaled by the configured image supervision weight.

\paragraph{Encoder.}
The encoder uses a bidirectional self-attention mask over all input tokens, including current visual tokens, action tokens, and learnable register tokens.
This design allows every token to access the entire action chunk and the visual environment.
In particular, the register tokens can aggregate temporal information across the full action sequence while interacting with environment context, so that the latent action tokens are formed from both executable action structure and scene-dependent information.

\paragraph{Decoder.}
The decoder uses an asymmetric attention mask tailored to the heterogeneous reconstruction targets.
For future visual feature prediction, visual queries are allowed to attend to the full decoder token set, including latent tokens and action-token states.
This enables the visual branch to predict the endpoint scene consequence from complete action and latent information.
For action reconstruction, action queries follow a causal reconstruction mask: each action position can attend to the latent action tokens and its available causal action context, but is prevented from attending to future action positions and visual prediction tokens.
This preserves the autoregressive structure of action reconstruction and prevents exploiting future visual tokens as a shortcut.
As a result, the decoder couples action reconstruction and visual consequence prediction through shared latent conditioning, while maintaining a clean generation path from action embedding tokens to executable controls.

\subsection{Training Pipeline}
The training adopts a two-stage pipeline. In the first stage, EDAR is trained to capture environment-dependent, continuous latent representations of action trajectories. In the second stage, the pretrained EDAR is embedded into a policy model, and this policy is trained for downstream control tasks while EDAR’s parameters are kept frozen. We use fixed random seeds to ensure reproducibility.

\begin{table*}[t]
\centering
\caption{\textbf{Training recipes for EDAR representation learning.}}
\label{tab:edar_stage1_training_recipes}
\small
\setlength{\tabcolsep}{7pt}
\renewcommand{\arraystretch}{1.12}
\begin{tabular}{lcccc}
\toprule
\multirow{2}{*}{Parameter} 
& \multicolumn{3}{c}{Simulation} 
& \multirow{2}{*}{Real-world} \\
\cmidrule(lr){2-4}
& LIBERO & CALVIN & Meta-World & \\
\midrule
Latent action dim 
& 7 & 7 & 4 & 7 \\
Image supervision weight $\lambda$ 
& 0.2 & 0.1 & 0.2 & 0.2 \\
\midrule
Training steps 
& \multicolumn{3}{c}{100k} & 100k \\
Batch size 
& \multicolumn{3}{c}{64} & 128 \\
Chunk size 
& \multicolumn{3}{c}{16} & 20 \\
Latent action tokens $K$ 
& \multicolumn{3}{c}{4} & 4 \\
Learning rate schedule 
& \multicolumn{3}{c}{$1e^{-4} \rightarrow 1e^{-5}$} 
& $2e^{-4} \rightarrow 2e^{-5}$ \\
Optimizer 
& \multicolumn{3}{c}{Adam} & Adam \\
Optimizer betas 
& \multicolumn{3}{c}{$(0.9,\ 0.99)$} 
& $(0.9,\ 0.99)$ \\
\bottomrule
\end{tabular}
\end{table*}

\paragraph{\textbf{LIBERO.}}
For LIBERO experiments, we use the cleaned and reformatted dataset released by the HuggingFace team following the preprocessing pipeline of OpenVLA~\cite{kim2024openvla}. The dataset removes no-op frames and failed episodes, and contains 1,693 expert demonstrations converted to the LeRobot format. We use the same dataset version as the SmolVLA baseline to ensure a fair comparison.

EDAR is first trained in a representation-learning stage using the training setting in Table~\ref{tab:edar_stage1_training_recipes}, with action reconstruction and future visual consequence prediction as the learning objectives. We then integrate the pretrained EDAR module into SmolVLA by replacing the original action interface with EDAR latent action tokens. The downstream policy is trained on the same LIBERO dataset for 100k steps with batch size 64 and chunk size 16. The learning rate follows a cosine decay schedule from $1.5\mathrm{e}{-4}$ to $5\mathrm{e}{-6}$. During evaluation, the policy predicts 16 future actions at each step and executes the first 8 actions.

\paragraph{\textbf{CALVIN.}}
For CALVIN, we use the official CALVIN ABCD$\rightarrow$D benchmark converted into the LeRobot format. CALVIN is a long-horizon language-conditioned manipulation benchmark built from human-teleoperated demonstrations in a tabletop environment~\cite{mees2022calvin}. The dataset contains four environment splits, denoted A, B, C, and D; in the standard ABCD$\rightarrow$D protocol, policies are trained on splits A--D and evaluated on the held-out split D. 

We first train EDAR on the CALVIN dataset using the recipe in Table~\ref{tab:edar_stage1_training_recipes}. The EDAR module is integrated into SmolVLA, and the downstream policy is trained on the same dataset for 100k steps with batch size 64 and chunk size 16. The learning rate follows a cosine schedule decaying from $1.5\mathrm{e}{-4}$ to $5\mathrm{e}{-6}$. During evaluation, the policy predicts 16 actions and executes the first 8 at each step.

We additionally integrate the same CALVIN-trained EDAR module into FLOWER-VLA~\cite{reuss2025flower}. For controlled comparison, EDAR only replaces the action representation interface, while all FLOWER-VLA training hyperparameters and evaluation settings are kept unchanged, including the training budget, batch size, optimizer and scheduler settings, denoising architecture, and evaluation action horizon. Following the original setup, the policy is trained for 40k steps across 4 GPUs with batch size 8 per GPU, and the action horizon is set to 10 during evaluation. Different from the original DDP implementation, our reproduction uses FSDP for distributed training.

\paragraph{\textbf{Meta-World.}}
For Meta-World, we follow the standard MT50 evaluation setting used in SmolVLA~\cite{shukor2025smolvla} and evaluate on 50 manipulation tasks grouped into Easy, Medium, Hard, and Very Hard difficulty categories.

We first train EDAR on Meta-World using the representation-learning recipe in Table~\ref{tab:edar_stage1_training_recipes}, with action reconstruction and future visual consequence prediction as the learning objectives. The original SmolVLA paper reports Meta-World results, but omits some Meta-World-specific hyperparameters needed for exact reproduction.
We therefore build a controlled SmolVLA-Meta-World baseline under our verified training recipe. The baseline loads the pretrained VLM weights and uses a four-dimensional state and action interface, chunk size 16, action horizon 8, 100k scheduler decay steps, and a cosine learning-rate schedule from $1.5\mathrm{e}{-4}$ to $5\mathrm{e}{-6}$. We then integrate the pretrained EDAR module into the same reproduced SmolVLA pipeline by replacing the original action interface with EDAR latent action tokens. All other settings, including data processing, training schedule, learning-rate schedule, evaluation code, and checkpoint selection, are kept identical. 

\paragraph{Real-World Experiments.}
For real-world EDAR representation learning, we adopt a two-stage training procedure. In the first stage, EDAR is pretrained on 10 Hz, 15 Hz, and 20 Hz subsets from the OpenX dataset~\cite{o2024open}, where samples are drawn uniformly across individual datasets. 
In the second stage, EDAR is further trained on our self-collected tele-operated real-world data. Both stages follow the same real-world training recipe in \cref{tab:edar_stage1_training_recipes}. 

For downstream policy learning, the pretrained EDAR module is integrated into SmolVLA by replacing the original action interface with EDAR latent action tokens. The downstream policy is trained on the same self-collected task data for 100k steps with batch size 64 and chunk size 20. The learning rate follows a cosine decay schedule from $1.5{\times}10^{-4}$ to $5{\times}10^{-6}$. The SmolVLA baseline uses the same policy training setup.

\section{Environment-dependent Structure of Action Latent Space}
\label{sec:environment_conditioned_latent_visualization}

\begin{figure*}[t]
\centering
\includegraphics[width=\linewidth]{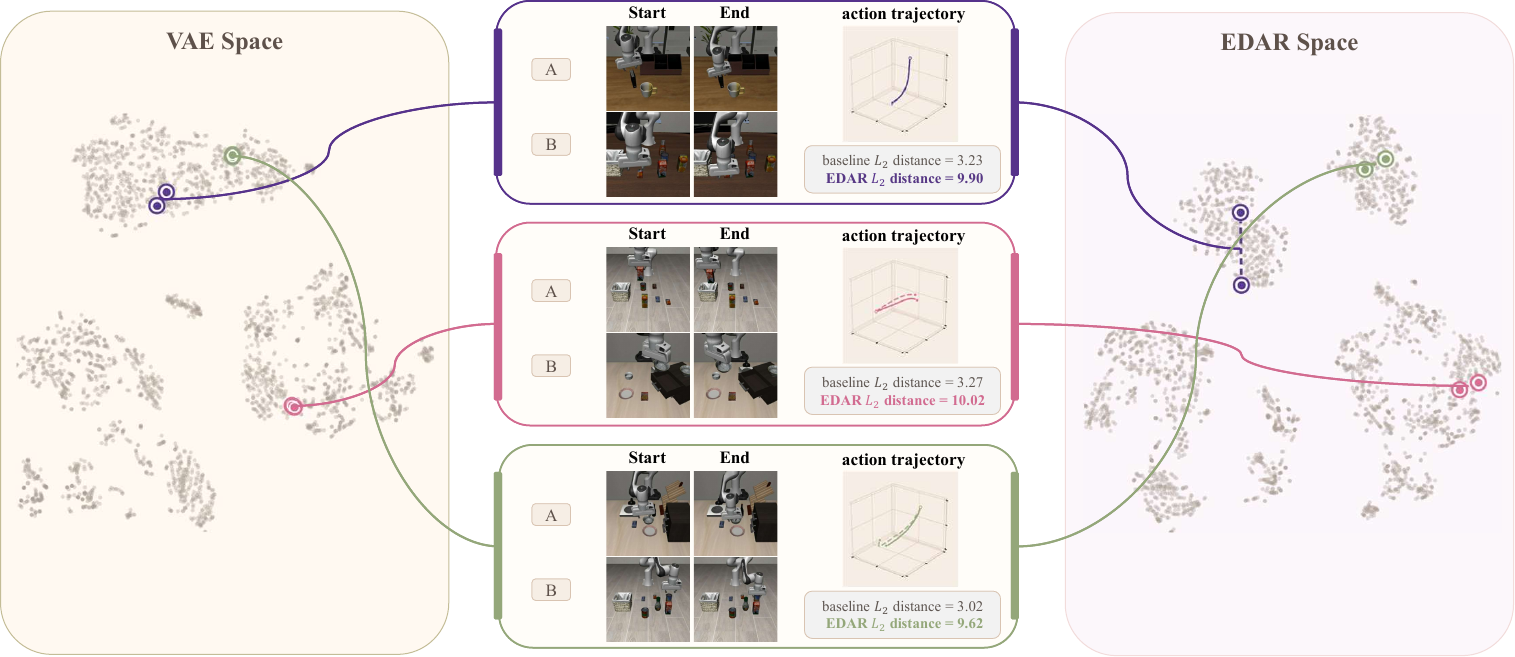}
\caption{
\textbf{Environment-dependent structure of action latent space.}
We compare the environment-independent VAE and EDAR on the same three matched action pairs from LIBERO. Each middle panel shows one pair from two different episodes, with start/end observations and an action trajectory indicating similar short-horizon command structure. In the environment-independent baseline latent map, each pair remains close, while in the EDAR latent map the same pair becomes more separated under different visual contexts. The reported \(L_2\) distances measure the separation between the two highlighted points in their raw latent spaces.}
\label{fig:joint_baseline_edar_contrast}
\end{figure*}

Figure~\ref{fig:joint_baseline_edar_contrast} illustrates the environment-dependent structure of the action latent space by comparing action chunks with similar command-level motion but different scene contexts. The goal is to examine whether EDAR assigns different latent representations to similar motor patterns when the environment changes their expected consequences.

We construct the visualization from 16-step action chunks in the LIBERO demonstration dataset. For each displayed pair, the two chunks have similar local command structure, as shown by the near-overlapping action trajectories, but are executed in different visual contexts. We compare an environment-independent action encoder with EDAR. The baseline encodes each action chunk without the current observation, while EDAR encodes the same chunk together with its corresponding current visual observation. We then project the latent representations into two dimensions with t-SNE. Gray points show sampled background chunks, and colored points highlight the matched pairs.

Each middle panel shows one matched pair from two LIBERO episodes. The start and end observations show the scene before and after the 16-step horizon, illustrating that similar motions can correspond to different environment-conditioned consequences. The reported pairwise distance is computed as the Euclidean distance between the two corresponding representations in the latent space.

Across the three examples, the environment-independent encoder places each pair almost on top of each other, reflecting their similar command-level motion. In contrast, EDAR separates the same pairs more clearly when they occur under different scene contexts. This suggests that EDAR does not merely encode short-horizon motor patterns, but allows scene context to modulate the latent action representation. Similar action chunks can therefore occupy different regions of the latent space when they are associated with different expected effects on the environment.

\section{Visualization of Visual Context Prediction}
\label{sec:visual_context_quality}

\begin{figure*}[t]
\centering

\begin{subfigure}{\linewidth}
    \centering
    \includegraphics[width=\linewidth]{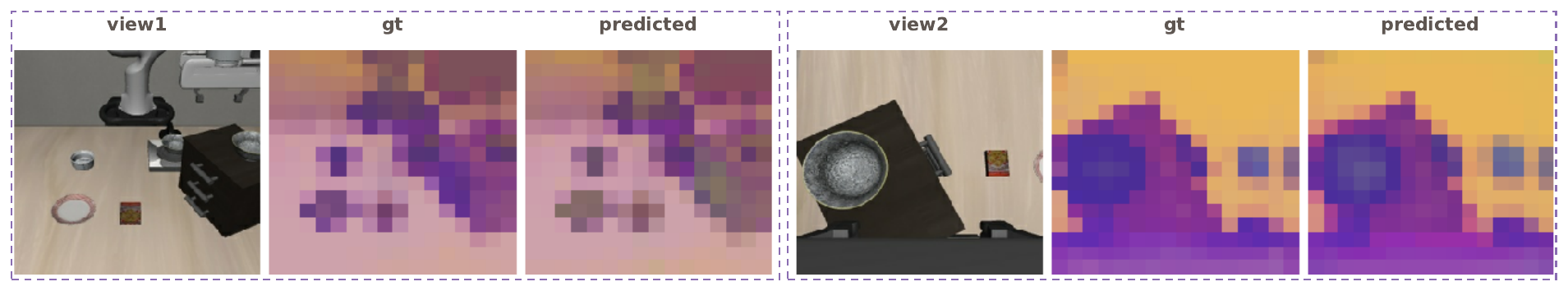}
    \caption{DINOv3.}
\end{subfigure}

\vspace{0.8em}

\begin{subfigure}{\linewidth}
    \centering
    \includegraphics[width=\linewidth]{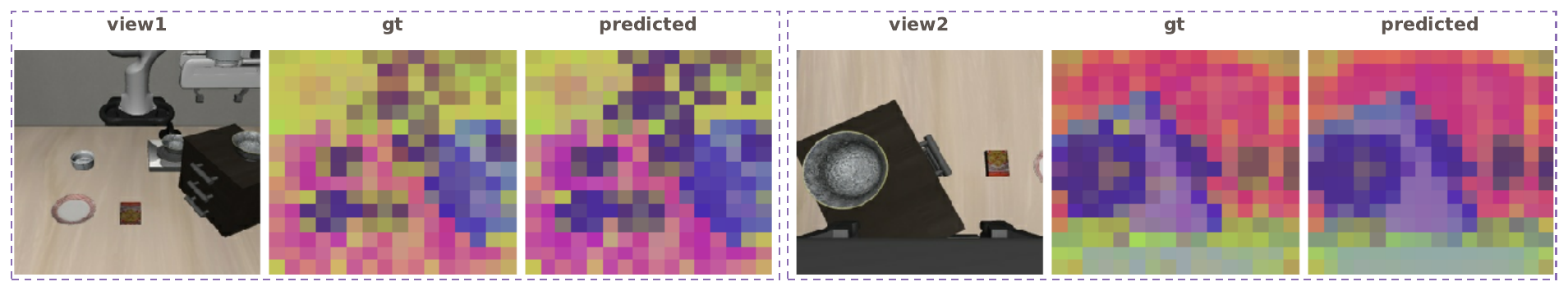}
    \caption{V-JEPA2.1.}
\end{subfigure}

\caption{\textbf{Visualization of visual context prediction.}
We compare the predicted future feature maps when EDAR is supervised with DINOv3 and V-JEPA2.1 features on the same black-bowl segment. Each subfigure corresponds to one target feature space. Within each subfigure, we show both the main-camera view and the wrist-camera view; for each view, the panels are ordered as the future RGB observation, ground-truth future feature map, and predicted future feature map. The colored maps are PCA projections of patch-level visual features in the corresponding encoder space.
}
\label{fig:visual_context_quality}
\end{figure*}

Figure~\ref{fig:visual_context_quality} visualizes future visual-context feature prediction for a randomly selected LIBERO action segment under the standard DINOv3-based EDAR checkpoint and a V-JEPA2.1 variant. Since the ground truth is a feature field rather than an RGB reconstruction, each example shows the raw future image together with palette-stylized PCA projections of the ground-truth and predicted future patch features.
The left and right sides of each strip correspond to the two camera views used as visual context.

The visualization helps explain why DINOv3 features serve as a stronger visual target for EDAR. Since EDAR predicts an endpoint future feature field rather than an RGB image or a full video rollout, the target feature space should provide spatially stable and locally interpretable supervision. In Fig.~\ref{fig:visual_context_quality}, the DINOv3 projections show clearer separation between task-relevant objects, robot parts, and background regions. The predicted DINOv3 feature field also preserves much of this region-level layout, suggesting that the auxiliary loss provides a clean signal for associating action latents with object-centric scene changes.

By contrast, the V-JEPA2.1 projections appear more locally fragmented, with weaker separation between foreground objects and background regions. Although V-JEPA2.1 is designed for image-video representation learning, its representation space is less aligned with EDAR's static endpoint dense-feature regression objective. This mismatch may make the visual consequence loss less informative for organizing action latents.

\section{Qualitative Results}
\label{sec:qualitative_results}

We provide qualitative visualizations of successful rollouts across all benchmarks. In each figure, every row shows one task rollout, with the leftmost frame depicting the initial state and the rightmost frame depicting the final state.

\subsection{Qualitative Results on LIBERO}

\begin{figure*}[t]
\centering

\begin{subfigure}{\linewidth}
    \centering
    \includegraphics[width=\linewidth]{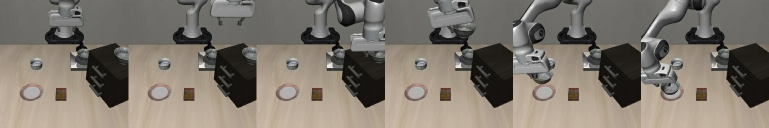}
    \caption{Spatial: pick up the black bowl on the wooden cabinet and place it on the plate.}
\end{subfigure}

\begin{subfigure}{\linewidth}
    \centering
    \includegraphics[width=\linewidth]{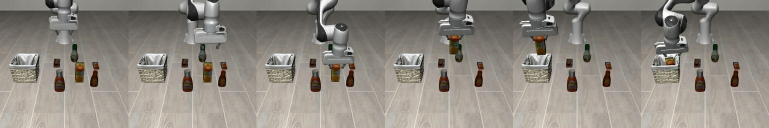}
    \caption{Object: pick up the orange juice and place it in the basket.}
\end{subfigure}

\begin{subfigure}{\linewidth}
    \centering
    \includegraphics[width=\linewidth]{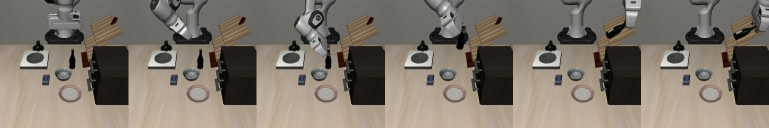}
    \caption{Goal: put the wine bottle on the rack.}
\end{subfigure}

\begin{subfigure}{\linewidth}
    \centering
    \includegraphics[width=\linewidth]{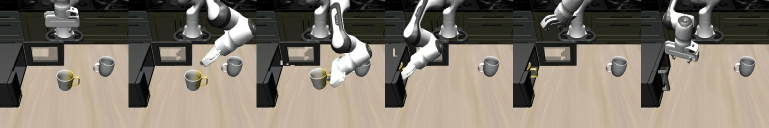}
    \caption{Long: put the yellow and white mug in the microwave and close it.}
\end{subfigure}

\caption{
\textbf{Qualitative results on LIBERO tasks.}
Representative examples include tasks from Spatial, Object, Goal, and Long.
}
\label{fig:libero_vis}
\end{figure*}

As illustrated in Figure~\ref{fig:libero_vis}, we show sampled rollout frames from the four LIBERO task suites: \emph{Spatial}, \emph{Object}, \emph{Goal}, and \emph{Long}. Each row corresponds to one task suite, where we uniformly sample six frames from one representative episode.

\subsection{Qualitative Results on CALVIN}

\begin{figure*}[t]
\centering

\begin{subfigure}{\linewidth}
    \centering
    \includegraphics[width=\linewidth]{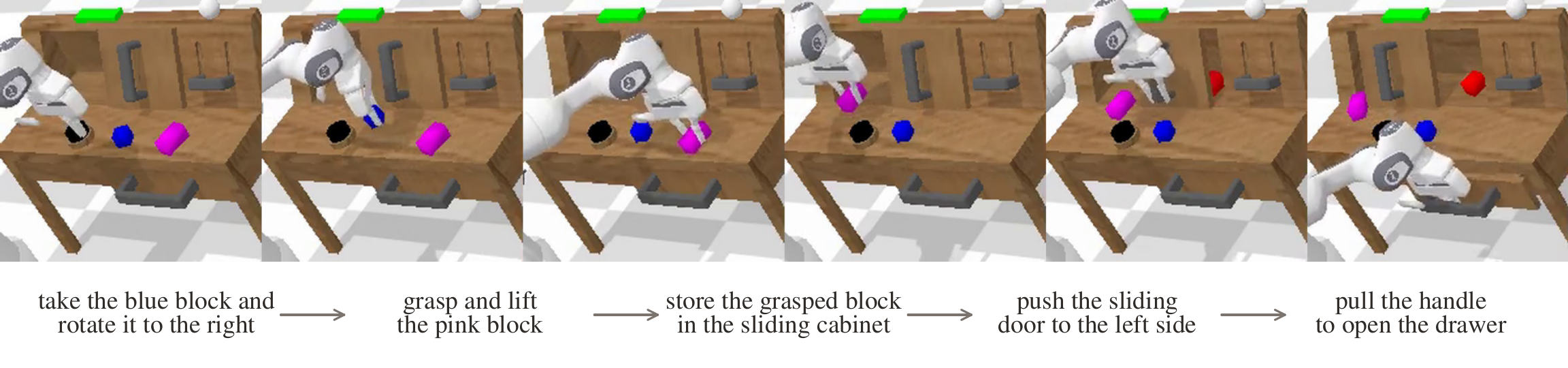}
\end{subfigure}

\begin{subfigure}{\linewidth}
    \centering
    \includegraphics[width=\linewidth]{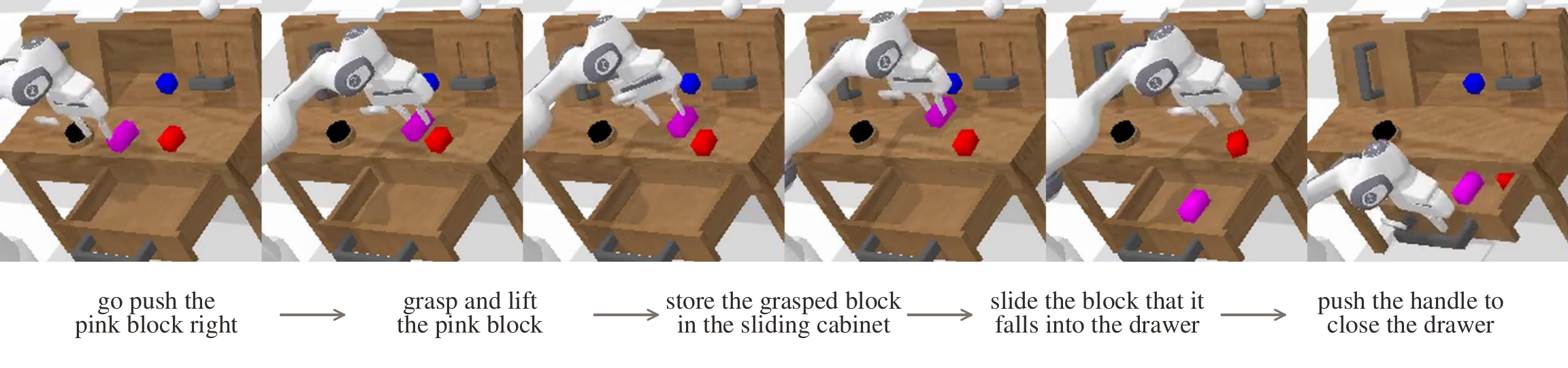}
\end{subfigure}

\begin{subfigure}{\linewidth}
    \centering
    \includegraphics[width=\linewidth]{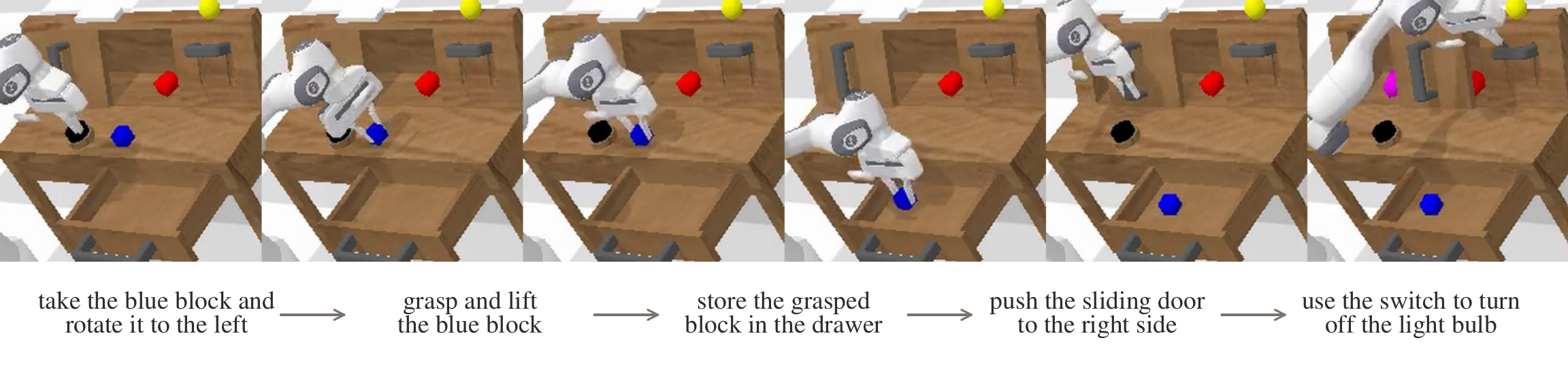}
\end{subfigure}

\begin{subfigure}{\linewidth}
    \centering
    \includegraphics[width=\linewidth]{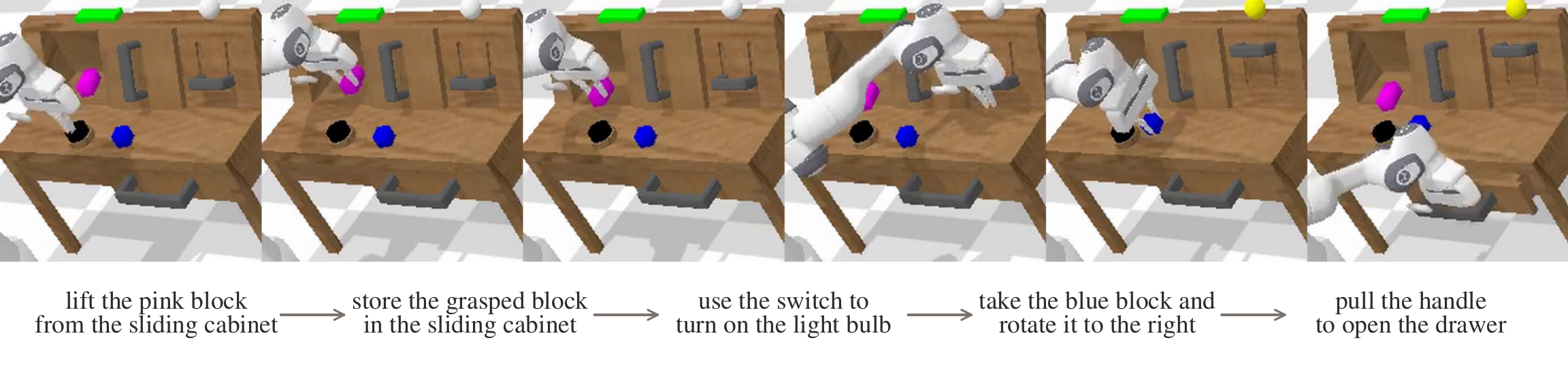}
\end{subfigure}

\begin{subfigure}{\linewidth}
    \centering
    \includegraphics[width=\linewidth]{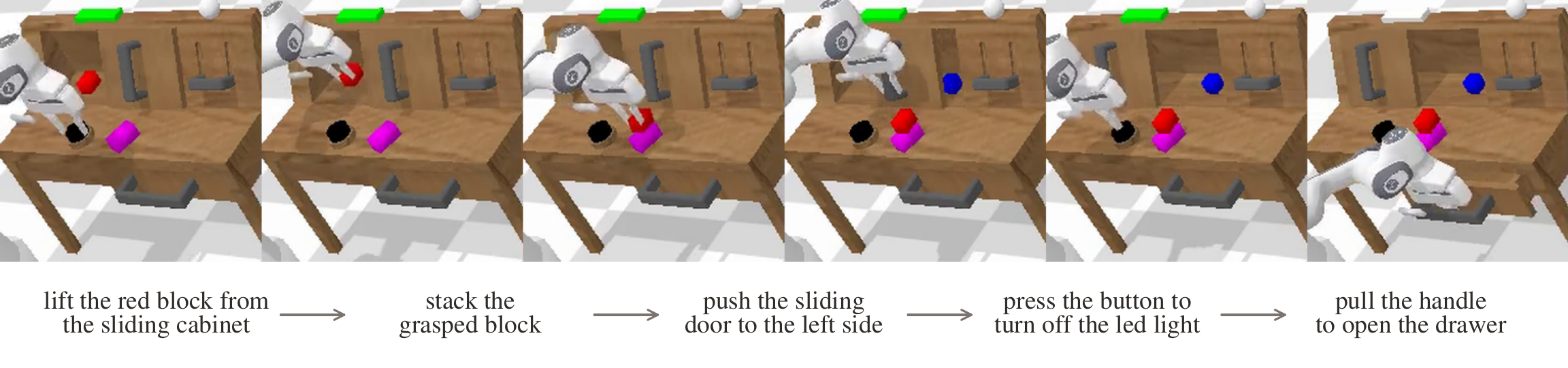}
\end{subfigure}

\caption{
\textbf{Qualitative results on CALVIN.}
Each row shows one successful 5-step task sequence reconstructed from the local CALVIN evaluation videos using only the main camera view.
}
\label{fig:calvin_vis}
\end{figure*}

Figure~\ref{fig:calvin_vis} visualizes successful multi-step CALVIN evaluation sequences. Each row is formed from six square main-view frames: the initial frame of the sequence followed by one late frame from each of the five successful subtasks in the same rollout sequence.

\subsection{Qualitative Results on Meta-World}

\begin{figure*}[t]
\centering

\begin{subfigure}{\linewidth}
    \centering
    \includegraphics[width=\linewidth]{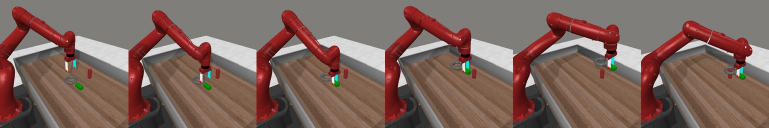}
    \caption{Pick up a nut and place it onto a peg.}
\end{subfigure}

\begin{subfigure}{\linewidth}
    \centering
    \includegraphics[width=\linewidth]{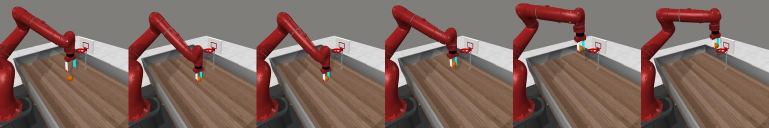}
    \caption{Dunk the basketball into the basket.}
\end{subfigure}

\begin{subfigure}{\linewidth}
    \centering
    \includegraphics[width=\linewidth]{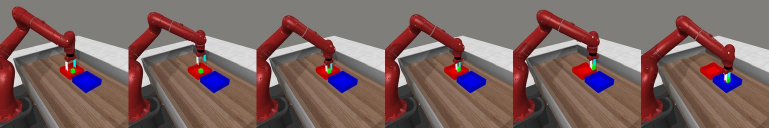}
    \caption{Grasp the puck from one bin and place it into another bin.}
\end{subfigure}

\begin{subfigure}{\linewidth}
    \centering
    \includegraphics[width=\linewidth]{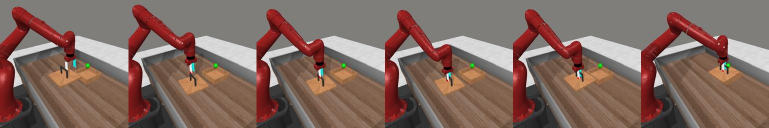}
    \caption{Grasp the cover and close the box with it.}
\end{subfigure}

\begin{subfigure}{\linewidth}
    \centering
    \includegraphics[width=\linewidth]{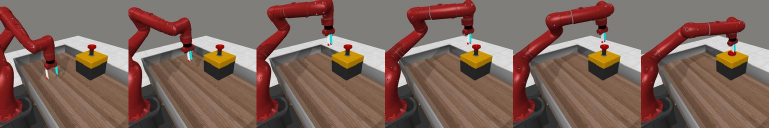}
    \caption{Press a button from the top.}
\end{subfigure}

\caption{
\textbf{Qualitative results on Meta-World.}
Representative examples include peg, basketball, puck-transfer, box-closing, and button-press tasks from the MT50 benchmark.
}
\label{fig:metaworld_vis}
\end{figure*}

Figure~\ref{fig:metaworld_vis} visualizes representative Meta-World MT50 tasks. Each row corresponds to one task and contains six uniformly sampled frames from one episode. The script uses the local cache when available and otherwise downloads only the required MT50 task shards.

\end{document}